\documentclass{article}

\usepackage[preprint]{spconfa4}
\usepackage{amsmath,epsfig}
\usepackage{amssymb,amsfonts}
\usepackage{multirow}
\let\OLDthebibliography\thebibliography
\renewcommand\thebibliography[1]{
  \OLDthebibliography{#1}
  \setlength{\parskip}{0pt}
  \setlength{\itemsep}{0pt plus 0.3ex}
}

\begin{document}\sloppy

\def\x{{\mathbf x}}
\def\L{{\cal L}}

\title{A Unified Two-Stage Group Semantics Propagation and Contrastive Learning Network for Co-Saliency Detection}
%
\name{Zhenshan Tan$^{\dagger}$, Cheng Chen$^{\dagger}$, Keyu Wen$^{\dagger}$, Yuzhuo Qin$^{\dagger}$ and Xiaodong Gu$^{\dagger,\ast}$\thanks{$^{\ast}$Corresponding Author. This work was supported in part by National Natural Science Foundation of China under grant 62176062.}}
\address{$^{\dagger}$Department of Electronic Engineering, Fudan University, Shanghai 200433, China \\xdgu@fudan.edu.cn}

\maketitle

\begin{abstract}
Co-saliency detection (CoSOD) aims at discovering the repetitive salient objects from multiple images. Two primary challenges are group semantics extraction and noise object suppression. In this paper, we present a unified Two-stage grOup semantics PropagatIon and Contrastive learning NETwork (TopicNet) for CoSOD. TopicNet can be decomposed into two substructures, including a two-stage group semantics propagation module (TGSP) to address the first challenge and a contrastive learning module (CLM) to address the second challenge. Concretely, for TGSP, we design an image-to-group propagation module (IGP) to capture the consensus representation of intra-group similar features and a group-to-pixel propagation module (GPP) to build the relevancy of consensus representation. For CLM, with the design of positive samples, the semantic consistency is enhanced. With the design of negative samples, the noise objects are suppressed. Experimental results on three prevailing benchmarks reveal that TopicNet outperforms other competitors in terms of various evaluation metrics.
\end{abstract}
\begin{keywords}
Co-saliency detection, group semantics extraction, noise object suppression, contrastive learning
\end{keywords}
\section{Introduction}
\label{sec:intro}

Co-salient object detection (CoSOD) not only imitates the human visual system like salient object detection (SOD) to detect salient objects, but also considers whether the salient objects from a group of images appear repeatedly. This means that CoSOD needs to discover the repetitive salient objects and suppress non-repetitive salient and non-salient ones. Benefiting from the above favorable characteristics, CoSOD is widely applied as a pre-processing step in various down-streaming tasks, such as RGB-D task \cite{cong2018hscs}, visual tracking \cite{li2018visual}, video surveillance \cite{gao2020trustful} and video retrieval \cite{hong2017coherent}.

Group semantics extraction, which explores the common features of multiple relevant objects, directly determines the resulting performances. It has two key points: 1) maximizing the intra-group co-salient object compactness and 2) minimizing the noise object interference. Recently, the end-to-end learning-based architectures \cite{fan2021group, zhang2020gradient, li2021image, fan2021re, ren2020co} achieve promising results. However, they still suffer from two issues in group semantics extraction. On the one hand, these architectures ignore the vital relevancy learning of group semantics, which decreases the compactness of co-salient objects. They extract group semantics with one-stage model and only focus on the common features such as classification consensus representation, merged convolutional features and intra-group common attributes. However, these methods lack the relevancy learning of these generated common features, which decreases the semantic distribution connection. This means that the learned group semantics are scattered and vulnerable, and it is easy to produce less robust performance in the inference. On the other hand, previous methods \cite{zhang2020gradient,li2021image,fan2021re,ren2020co} ignore the distinction between categories or completely negate the correlation between categories \cite{fan2021group}, which fails to suppress noise objects. The former pays more attentions to the positive relations of intra-group repetitive objects and lacks the learning of the negative relations among inter-group objects, which leads to the lack of outlier object detection. The latter suppresses the negative relationship to zero to distinguish the inter-group objects. However, the features between objects with similar but different classes usually have overlapping parts. Completely denying the correlation between categories may mislead semantic learning.

With the above insights, we propose an end-to-end unified Two-stage grOup semantics PropagatIon and Contrastive learning NETwork (TopicNet). To address the first issue, a two-stage group semantics propagation mechanism (TGSP) is presented. In the first stage, we design an image-to-group propagation module (IGP). By the aid of IGP, the similar features of the relevant objects with same category are propagated within the group. To build the semantic relevancy, a group-to-pixel propagation module (GPP) is embedded after IGP to calculate the semantics similarity among pixels. To address the second issue, the contrastive learning is introduced to CoSOD. The core point of the second issue is to pull the group semantics to the adjacent features to enhance the semantic consistency and push away to the noise features. Inspired by recent work \cite{chen2020simple} of contrastive learning that pulls the positive samples closer to the anchor and pushes the negative samples farther away, we naturally apply the contrastive learning to CoSOD by reasonably designing positive and negative samples.  Besides, compared with the method \cite{fan2021group}, we also maximize the separability between different categories but we do not completely deny the inter-class correlation, which leaves room for the semantic learning of similar categories. In addition, TopicNet can conveniently produce more interference features and add more groups to enhance the network generalization ability. Note that contrastive learning is optimized only during training, which does not impose an extra burden to the inference. To the best of our knowledge, this is the first attempt to introduce contrastive learning into CoSOD.

In summary, the main contributions of this paper can be concluded as follows.

\begin{itemize}
  \item We employ a two-stage group semantics propagation mechanism (TGSP), including an image-to-group propagation module (IGP) and a group-to-pixel propagation module (GPP). IGP propagates intra-group similarity features to explore group semantics and GPP distributes the group semantics to each pixel, which maximizes the intra-group co-salient object compactness.
  \item We introduce contrastive learning to CoSOD. With the design of positive samples and negative samples, the intra-group semantic representation is enhanced, the different categories are distinguished and the noise objects are suppressed.
  \item Experimental evaluations demonstrate the proposed end-to-end network outperforms previous competitors.
\end{itemize}

\section{Related Works}
In this section, we briefly review some related methods of CoSOD and more related works can be referenced by the literature \cite{fan2021re}.

Early methods rely on various prior cues such as color and texture. However, these hand-crafted features still far away from being satisfactory in complex scenes due to the discriminative limitation. With the great progress of deep neural networks (DNNs) based methods in various fields, some researchers attempt to combine the prior pre-processing model with DNNs \cite{zhang2018co, jiang2019multiple, zhang2015self, zhang2016detection, han2017unified, zhang2019co}. Nevertheless, the results of these methods depend in part on the prior features and the pre-processing module brings extra computing overhead and computing time.

Recent deep-based methods attempt to detect co-salient objects with an end-to-end manner. Li et al. \cite{li2021image} introduce an attention graph clustering model to graph neural networks to explore the group semantics. Zhang et al. \cite{zhang2020gradient} extract group semantics by the pre-training classification features and calibrate the individuals with feedback gradient information. Fan et al. \cite{fan2021re} design a class activation mapping and principal component analysis to learn co-saliency features. Ren et al. \cite{ren2020co} concatenate multi-layer features and then explore group semantics with feature aggregation. Zhang et al. \cite{zhang2020coadnet} propose an aggregate-distribute architecture to explore the co-attention and a group consistency decoder to keep the constraints. Different from previous methods, Fan et al. \cite{fan2021group} learn the negative relations with adding another image group to maximize inter-class separability. Although the above methods improve the results, they ignore the relevancy learning of group semantics and the noise object suppression, which are to be addressed in this paper.

\begin{figure*}
	\centering
	\includegraphics[width=\textwidth]{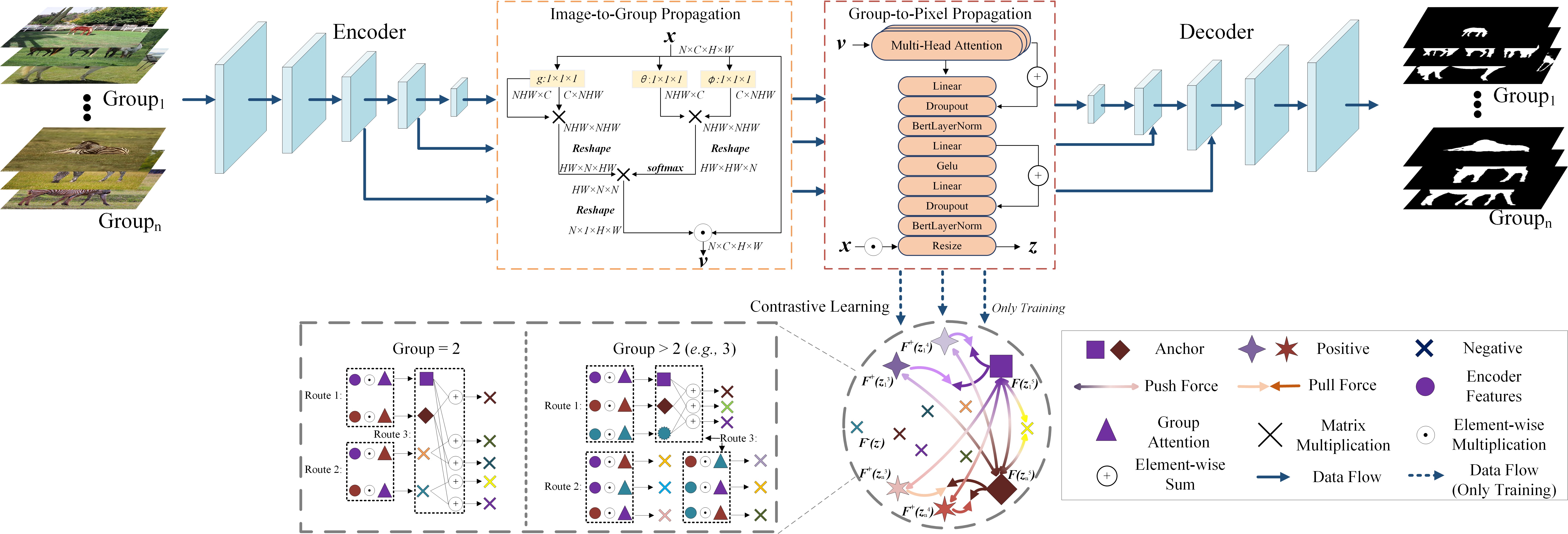}
	\caption{Pipeline of the proposed unified Two-stage grOup semantics PropagatIon and Contrastive learning NETwork (TopicNet).}
	\label{fig2}
\end{figure*}

\section{Proposed Method}
\subsection{Architecture Overview}
Fig.~\ref{fig2} illustrates the flowchart of the unified Two-stage grOup semantics PropagatIon and Contrastive learning NETwork (TopicNet). Given $M$ groups with $N$ images in each group $\{I_{1,n},I_{2,n},...,I_{M,n}\}_{n=1}^N$, firstly, the last three layers of a weight-shared encoder are adopted to generate feature map $\{x_{m,n}^i\}_{n=1}^N\in \mathbb{R}^{N\times C\times H\times W}$ for $M$ groups, where $i=3,4,5$ represents the layer serial number, $C$ is the channel number and $H\times W$ is the dimensional size. Secondly, $\{x_{m,n}^i\}_{n=1}^N$ is fed into the image-to-group propagation module (IGP) to preliminarily extract group semantics $\{r_{m,n}^i\}_{n=1}^N\in \mathbb{R}^{N\times C\times H\times W}$. Thirdly, $\{r_{m,n}^i\}_{n=1}^N$ is transferred to group-to-pixel propagation module (GPP) to distill the co-attention $\{e_{m,n}^i\}_{n=1}^N\in \mathbb{R}^{1\times C\times 1\times 1}$. Then, $\{x_{m,n}^i\}_{n=1}^N$ element-wise multiplied by $\{e_{m,n}^i\}_{n=1}^N$ explores the recalibrated group semantics $\{z_{m,n}^i\}_{n=1}^N\in \mathbb{R}^{N\times C\times H\times W}$. Fourthly, $\{z_{m,n}^i\}_{n=1}^N$ is fed into the contrastive learning module to enhance semantic consistency and suppress noise objects. Finally, the refined group semantics are aggregated into the decoder to modify the individuals to predict the salient maps $\{\mathcal{M}_{m,n}\}_{n=1}^N$.

\subsection{Two-Stage Group Semantics Propagation}
\subsubsection{Image-to-Group Propagation}
For multiple images in a group, the co-salient objects share similar features. Inspired by non-local neural network (NLN) \cite{wang2018non}, we propose an IGP to explore these similar features to calculate the global consensus representation. As illustrated in Fig.~\ref{fig2}, different from the classic NLN that propagates the inter-pixel corresponding relationship of an image, IGP focuses on improving the inter-image correspondences. 

Firstly, $\{x_{m,n}^i\}_{n=1}^N$ coming from the encoder is resized to $14\times 14$ and then fed into IGP, aiming at saving the video memory. Without losing generality, the group subscript, the image subscript and the layer superscript are dropped. Secondly, the resized $x$ goes through three $1\times 1\times 1$ convolutional layers and the corresponding similarity weights $A_g\in \mathbb{R}^{NHW\times NHW}$ and $A_{\theta\phi}\in \mathbb{R}^{NHW\times NHW}$ are obtained. Then, $A_g$ and $A_{\theta\phi}$ are reshaped to $HW\times HW\times N \times N$. After that, we maximize the last dimensional $N$ to induce the features $A_g\in \mathbb{R}^{HW\times HW\times N}$ and $A_{\theta\phi}\in \mathbb{R}^{HW\times HW\times N}$ that have the greatest impact on the group semantics between images. Thirdly, $A_g$ is reshaped to $HW\times N\times HW$ and then matrix multiplies the normalized similarity weight $A_{\theta\phi}$, aiming at inducing the similarity tensors $A\in \mathbb{R}^{HW\times N\times N}$ between images. Then $A$ is averaged to calculate the mean of all the most influential features, generating the inter-image global attention. Finally, $A$ is normalized by softmax and multiplied by $x$ to preliminarily extract group semantics $r$.

\subsubsection{Group-to-Pixel Propagation}
Previous methods stop at the image-to-group stage. However, the group semantics in this stage lack the correlations between pixels and group semantics. To address this, we further propose a GPP to distribute group semantics to pixels. Structurally, a self-attention mechanism \cite{wen2020learning} is adopted in GPP. Then the distilled co-attention $\{e_{m,n}^i\}_{n=1}^N$ element-wise multiplies $\{x_{m,n}^i\}_{n=1}^N$ to generate group semantics $\{z_{m,n}^i\}_{n=1}^N$. As shown in Fig.~\ref{fig2}, different from IGP, this self-attention mechanism recalculates the correlations between pixels. Obviously, the consensus representation in IGP lacks sufficient semantic connection, suppressing some subtle features during propagation. GPP enhances relevancy learning of consensus representation by calculating the similarity between pixels, so as to realize the recalibration of group semantics.

\begin{figure*}
	\centering
	\includegraphics[width=\textwidth]{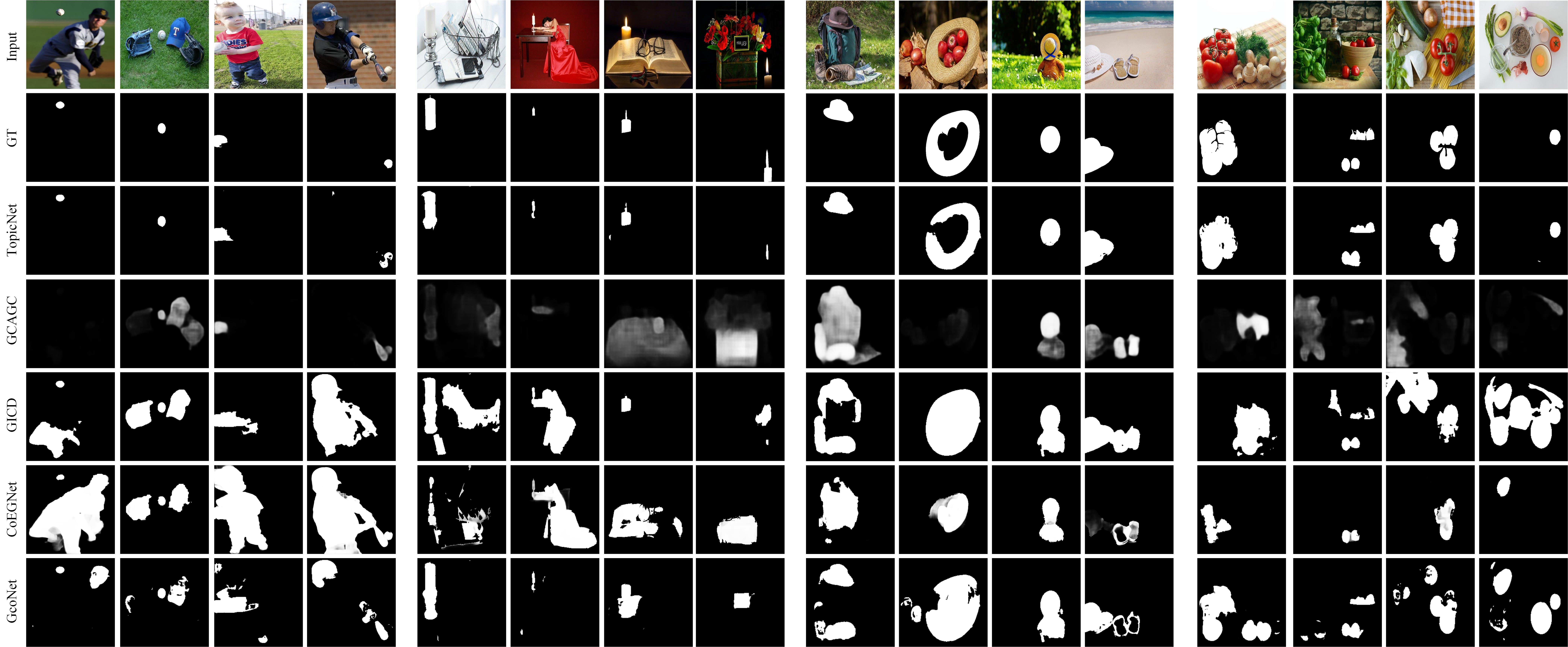}
	\caption{Qualitative comparisons of TopicNet and other competitors.}
	\label{fig3}
\end{figure*}

\subsection{Contrastive Learning}
Contrastive learning aims to pull close the positive samples and push away the negative samples at the same time. As mentioned in section 1, reasonably designing positive and negative samples can intuitively enhance the semantic consistency and suppress noise objects. Concretely, the relationship $\Psi^+$ between positive samples and the anchor is defined as.
\begin{equation}\label{eq3.3.1}
    \begin{split}
        F(z)&=x_{m,n}^5\odot \{e_{m,n}^5\}_{n=1}^N,\\\
        F(z^+)&=x_{m,n}^i\odot \{e_{m,n}^i\}_{n=1}^N,i=3,4,\\
        \Psi^+(F(z),F(z^+))&=exp(\frac{{F(z)}^TF(z^+)}{\tau||F(z)||||F(z^+)||}),
\end{split}
\end{equation}
where, $F(z)$ is the anchor and $F(z^+)$ is the positive sample. The temperature parameter $\tau$ relaxes the element-wise multiplication and $\tau=0.07$ following previous method \cite{chen2020simple}. Why is it designed like this? Evidence \cite{sun2019deeply} suggests that as the complexities of network depth, blocks and topology increase, the risk of insufficient representation learning is posed. An intuitive approach is to supervise the network. However, the encoder supervision will lead to performance degradation. Therefore, many researches \cite{fan2021group, zhang2020gradient} add auxiliary supervision in the decoder. Unfortunately, CoSOD needs to calibrate the group semantics with encoder, which contradicts the previous supervision design. To alleviate this contradiction, we pull close to the last three layers to enhance the semantic consistency. Meanwhile, the unsupervised refined encoder will not reduce the generalization ability of the network. Why only the last three layers? Because the first two layers are too close to the input, pulling close to all layers will lead to the loss of key information in the shallow layer or the addition of useless information in the deep layer.

Fig.~\ref{fig2} lists three generation routes of negative samples. The first one $\Psi_1^-$ is that the inter-group objects are negative samples of each other, aiming to maximize the inter-group object separability.
\begin{equation}\label{eq3.3.2}
     \Psi_1^-(F(z),F_1^l(z^-))=exp(\frac{{F(z)}^TF_1^l(z^-)}{\tau||F(z)||||F_1^l(z^-)||}),
\end{equation}
where, $l$ is another group anchor. The second one $\Psi_2^-$ is that the element-wise multiplication of non-corresponding $x_{j,n}^i$ and $e_{k,n}^i$ $(j\neq k)$ is defined as negative samples.
\begin{equation}\label{eq3.3.3}
    \begin{split}
        F_2(z^-)&=x_{j,n}^i\odot e_{k,n}^i,j\neq k,\\
        \Psi_2^-(F(z),F_2(z^-))&=\sum_{j=1}^{M}\sum_{k=1}^{M}exp(\frac{{F(z)}^TF_2(z^-)}{\tau||F(z)||||F_2(z^-)||}).
\end{split}
\end{equation}

Different group attentions are equivalent to assigning the main attributes of other objects to the current object, which enhances the discriminative representations of the network. The last one $\Psi_3^-$ is the fusion of different attentions.
\begin{equation}\label{eq3.3.4}
    \begin{split}
        F_3(z^-)&=x_{j,n}^i\odot (e_{j,n}^i+e_{k,n}^i),j\neq k,\\
        \Psi_3^-(F(z),F_3(z^-))&=\sum_{j=1}^{M}\sum_{k=1}^{M}exp(\frac{{F(z)}^TF_3(z^-)}{\tau||F(z)||||F_3(z^-)||}).
\end{split}
\end{equation}

The fused attention represents aggregating the features of other groups semantics to the current group semantics, aiming at producing interference features to enhance the generalization ability of the network. Then the relationship $\Psi^-$ between the whole negative samples and the anchor can be summed as.
\begin{equation}\label{eq3.3.5}
     \Psi^-(F(z),F(z^-))=\sum_{h=1}^{H}\Psi_h^-(F(z),F_h(z^-)),
\end{equation}
where, if the number of groups is 2, then $H=3$, otherwise $H=2$ because negative samples are enough to support the learning of noise objects. Then the loss function of contrastive learning can be defined as.
\begin{equation}\label{eq3.3.6}
     \mathcal{L}_{cl}=-log(\frac{\Psi^+(F(z),F(z^+))}{\Psi^+(F(z),F(z^+))+\Psi^-(F(z),F(z^-))}).
\end{equation}

Finally, the optimized positive samples and anchor are integrated into the decoder to calibrate individuals with saliency supervision. The saliency loss function is as follows.
\begin{equation}\label{eq3.3.7}
    \mathcal{L}_s=1-\frac{\sum_{(p,q)}{\mathcal{M}(p,q)\mathcal{T}(p,q)}}{\sum_{(p,q)}{(\mathcal{M}(p,q)+\mathcal{T}(p,q))}},
\end{equation}
where, $\mathcal{M}$ is the predicted saliency map and $\mathcal{T}$ is the ground truth, $p$ and $q$ are the pixel coordinates. During training, the network is jointly optimized by $\mathcal{L}_{cl}$ and $\mathcal{L}_s$.
\begin{equation}\label{eq3.3.8}
    \mathcal{L}=\lambda_1\mathcal{L}_{cl}+\lambda_2\mathcal{L}_s,
\end{equation}
where, $\lambda_1$ and $\lambda_2$ are the parameters that balance the loss weights. We set $\lambda_1$ and $\lambda_2$ to 1 because they are equally important.

\begin{table*}[t]
	\caption{Quantitative Comparisons with Other Methods. `-' denotes no reports; `$\uparrow$' (`$\downarrow$') denotes the larger (lower) the value, the better the results; Values with bold and underline indicate the best and the suboptimal performance; `RI' denotes relative improvement.}
	\begin{center}
		\resizebox{\textwidth}{!}{\begin{tabular}{|c|c|ccccc|ccccc|ccccc|}
				\hline
				\multirow{2}{*}{Methods}&\multirow{2}{*}{Publish}&\multicolumn{5}{c|}{CoSal2015}&\multicolumn{5}{c|}{CoCA} &\multicolumn{5}{c|}{CoSOD3k}\\
				\cline{3-17}
				&&$F_\mu$$\uparrow$&$\epsilon$$\downarrow$&$F_\gamma$$\uparrow$&$E_\mu$$\uparrow$&$S_\alpha$$\uparrow$&$F_\mu$$\uparrow$&$\epsilon$$\downarrow$&$F_\gamma$$\uparrow$&$E_\mu$$\uparrow$&$S_\alpha$$\uparrow$&$F_\mu$$\uparrow$&$\epsilon$$\downarrow$&$F_\gamma$$\uparrow$&$E_\mu$$\uparrow$&$S_\alpha$$\uparrow$\\
				\hline
				MIL \cite{zhang2015self}&ICCV 2015&0.677&0.209&0.611&0.722&0.676&-&-&-&-&-&-&-&-&-&-\\
				CoDW \cite{zhang2016detection}&IJCV 2016&0.725&0.274&0.560&0.752&0.650&-&-&-&-&-&-&-&-&-&-\\
				UMLF \cite{han2017unified}&TCSVT 2017&0.730&0.269&0.542&0.772&0.665&-&-&-&-&-&0.689&0.277&0.529&0.768&0.641\\
				GoNet \cite{hsu2018unsupervised}&ECCV 2018&0.781&0.159&0.692&0.806&0.754&-&-&-&-&-&-&-&-&-&-\\
				CSMG \cite{zhang2019co}&CVPR 2019&0.834&0.131&0.747&0.843&0.775&0.539&0.127&0.478&0.712&0.608&0.764&0.148&0.680&0.824&0.712\\
				IML \cite{ren2020co}&NEUCOM 2020&0.712&0.155&0.620&0.788&0.736&-&-&-&-&-&-&-&-&-&-\\
				GICD \cite{zhang2020gradient}&ECCV 2020&0.867&0.071&0.834&0.886&0.843&0.547&0.126&0.503&0.714&0.657&0.799&0.079&0.763&0.848&0.797\\
				GCAGC \cite{li2021image}&TMM 2021&0.854&0.090&0.767&0.881&0.810&0.555&0.113&0.492&0.715&0.649&0.806&0.092&0.730&0.850&0.785\\
				CoEGNet \cite{fan2021re}&TPAMI 2021&0.868&0.078&0.833&0.883&0.838&0.549&0.106&0.493&0.713&0.610&0.798&0.084&0.757&0.837&0.777\\
				GCoNet \cite{fan2021re}&CVPR 2021&0.872&\underline{0.068}&0.836&0.887&0.845&0.580&0.105&0.528&\textbf{0.759}&0.672&0.806&\underline{0.071}&0.769&0.860&0.802\\
				TopicNet-2&&\underline{0.873}&\textbf{0.060}&\underline{0.841}&\underline{0.899}&\underline{0.854}&\underline{0.595}&\underline{0.100}&\textbf{0.545}&\underline{0.743}&\underline{0.683}&\underline{0.807}&\underline{0.071}&\textbf{0.772}&\textbf{0.864}&\textbf{0.808}\\
				RI (\%)&&+0.1&+11.8&+0.5&+1.4&+1.1&+2.6&+4.8&+3.2&-2.1&+1.6&+0.1&0&+0.4&+0.5&+0.7\\
				TopicNet-3&&\textbf{0.884}&\textbf{0.060}&\textbf{0.849}&\textbf{0.901}&\textbf{0.858}&\textbf{0.598}&\textbf{0.094}&\underline{0.544}&\underline{0.743}&\textbf{0.687}&\textbf{0.811}&\textbf{0.070}&\textbf{0.772}&\underline{0.861}&\underline{0.806}\\
				RI (\%)&&+1.4&+11.8&+1.6&+1.6&+1.5&+3.1&+10.5&+3.0&-2.1&+2.2&+0.6&+1.4&+0.4&+0.1&+0.5\\
				\hline
		\end{tabular}}
		\label{tab1}
	\end{center}
\end{table*}

\section{EXPERIMENTS}
\subsection{Evaluation Datasets and Criteria}
Following pervious methods \cite{fan2021group, zhang2020gradient}, the proposed TopicNet is evaluated on three prevailing datasets, including CoCA \cite{zhang2020gradient}, CoSal2015 \cite{zhang2016detection} and CoSOD3k \cite{fan2021re}. CoCA has 80 categories with 1295 images. It is the most challenging dataset that contains at least one noise object in each group. CoSal2015 is also a challenging dataset that consists of 50 groups with 2015 images. CoSOD3k is widely used in CoSOD and contains 160 groups with 3316 images in total. Five evaluation criteria, including maximum F-measure ($F_\mu$), mean absolute error ($\epsilon$), average F-measure ($F_\gamma$), maximum E-measure ($E_\mu$) and S-measure ($S_\alpha$), are adopted to evaluate methods.

\subsection{Implementation Details}
The proposed framework is implemented on Pytorch platform. The number of groups is set to 2 and 3 (TopicNet-2 and TopicNet-3) to quantitatively evaluate the results. TopicNet-2 and TopicNet-3 are equipped with a V100 GPU. Following previous researches \cite{fan2021group, zhang2020gradient}, TopicNet is trained end-to-end on the dataset \cite{ zhang2020gradient} and the data augmentation skills such as normalization, random horizontal flipping and random rotation are adopted to decrease over-fitting optimization. Besides, the classic vgg16 and feature pyramid network are adopted as the baseline. During training, all the inputs are resized as $224\times 224$. Adam optimizer with default parameters is applied. The learning rate is set to 1e-4 during 100 epochs and the batch size is set to 16.

\subsection{Competing Methods}
We compare the proposed TopicNet against the CoSOD methods with released codes. In visual comparison, only the representative TopicNet-2 is visualized for saving room.

Tab.~\ref{tab1} reports the performance statistics and the relative improvements between TopicNet and other methods. Obviously, our method outperforms other competitors across most datasets in terms of five evaluation metrics. It is worth mentioning that TopicNet-3 achieves better performance than TopicNet-2. However, TopicNet-3 takes up more video memories (16 G vs 9 G) during training. Note that more input groups only occupy more video memories during training, and their consumptions in inference are the same. Besides, our method can achieve real-time speed (around 50 fps), which greatly benefits various down-streaming methods. The visualized results in Fig.~\ref{fig3} show that TopicNet has superior visual performance, which verifies the effectiveness of the proposed method.

\begin{table}[t]
		\caption{Ablation Analyses of Various Structures. GAM is the module of method [28].}
		\begin{center}
			\begin{tabular}{|ccccc|ccc|}
				\hline
				\multicolumn{5}{|c|}{Module}&\multicolumn{3}{c|}{CoCA}\\
				\hline
				&GAM&IGP&GPP&CLM&$F_\mu\uparrow$&$\epsilon\downarrow$&$S_\alpha\uparrow$\\
				\hline
				(a)&$\checkmark$&&&&.524&.150&.635\\
				(b)&&$\checkmark$&&&.532&.140&.640\\
				(c)&$\checkmark$&&$\checkmark$&&.541&.130&.651\\
				(d)&&$\checkmark$&$\checkmark$&&.553&.126&.659\\
				(e)&$\checkmark$&&&$\checkmark$&.581&.111&.670\\
				(f)&&$\checkmark$&&$\checkmark$&.583&.105&.675\\
				(g)&$\checkmark$&&$\checkmark$&$\checkmark$&.574&.103&.671\\
				(h)&&$\checkmark$&$\checkmark$&$\checkmark$&.595&.100&.683\\
				\hline
			\end{tabular}
			\label{tab2}
		\end{center}
	\end{table}

\subsection{Ablation Studies}

The effective experiments of each proposed module are shown in Tab.~\ref{tab2}. Besides, we add the module GAM of recent method \cite{fan2021group} as control group to verify the effectiveness. For a fair comparison, we design the ablation experiments with only TopicNet-2. We can observe that compared with GAM \cite{fan2021group}, the proposed IGP achieves more competitive results. The control groups (e-h) show that contrastive learning can effectively boost the performance. Ablation studies verify that the proposed each module makes the network learn more discriminative semantics representations.
Besides, Tab.~\ref{tab3} reports the results of picking different layers. The last three layers can better maintain the semantic consistency. If all the layers are selected, the performances will be significantly reduced. Because the first two layers are close to the image, pulling close to these layers will lead to the loss of key features or the addition of useless features.

\begin{table}[t]
	\caption{Ablation Analyses of Various Positive Sample Design.}
	\begin{center}
		\begin{tabular}{|c|ccccc|}
			\hline
			\multirow{2}{*}{Layers}&\multicolumn{5}{c|}{CoSal2015}\\
			\cline{2-6}
			&$F_\mu$$\uparrow$&$\epsilon$$\downarrow$&$F_\gamma$$\uparrow$&$E_\mu$$\uparrow$&$S_\alpha$$\uparrow$\\
			\hline
			5&0.868&0.066&0.838&0.891&0.845\\
			4+5&0.867&0.064&0.831&0.894&0.849\\
			3+4+5&0.873&0.060&0.841&0.899&0.854\\
			2+3+4+5&0.875&0.069&0.843&0.887&0.844\\
			1+2+3+4+5&0.856&0.072&0.821&0.883&0.832\\
			\hline
		\end{tabular}
		\label{tab3}
	\end{center}
\end{table}

\section{CONCLUSION}
In this paper, we present a unified Two-stage grOup semantics PropagatIon and Contrastive learning NETwork (TopicNet) for co-saliency detection. Specifically, we employ an image-to-group propagation module to explore the consensus representation of intra-group similar features. Then, a group-to-pixel propagation module is designed to enhance relevancy learning of consensus representation to recalibrate group semantics. Besides, with the design of contrastive learning, the semantic consistency is enhanced and the noise objects are suppressed. Extensive comparative experiments and ablation studies verify the effectiveness of TopicNet and each proposed module. In addition, TopicNet can detect results in real time (around 50 fps on a single GPU). 


\end{document}